# The Relationship Between AND/OR Search and Variable Elimination

**Robert Mateescu** and **Rina Dechter**
School of Information and Computer Science
University of California, Irvine, CA 92697-3425
{mateescu,dechter}@ics.uci.edu

## Abstract

In this paper we compare search and inference in graphical models through the new framework of AND/OR search. Specifically, we compare Variable Elimination (VE) and memory-intensive AND/OR Search (AO) and place algorithms such as graph-based backjumping and no-good and good learning, as well as *Recursive Conditioning* [7] and *Value Elimination* [2] within the AND/OR search framework.

## 1 INTRODUCTION

It is convenient to classify algorithms that solve reasoning problems of graphical models as either search (*e.g.*, depth first, branch and bound) or inference (*e.g.*, variable elimination, join-tree clustering). Search is time exponential in the number of variables, yet it can be accomplished in linear memory. Inference exploits the model's graph structure and can be accomplished in time and space exponential in the problem's *tree-width*. When the tree-width is big, inference must be augmented with search to reduce the memory requirements. In the past three decades search methods were enhanced with structure exploiting methods. These improvements often require substantial memory, making the distinction between search and inference fuzzy. Recently, claims regarding the superiority of memory-intensive search over inference or vice-versa are made [3]. Our aim is to clarify this relationship and to create cross-fertilization using the strengths of both schemes.

In this paper we compare search and inference in graphical models through the new framework of AND/OR search, recently introduced [11]. Specifically, we compare Variable Elimination (VE) against memory-intensive AND/OR Search (AO), and place algorithms such as graph-based backjumping, no-good and good learning, and look-ahead schemes [9], as well as Recursive Conditioning [7] and Value Elimination [2] within the AND/OR search framework. We show that there is no principled difference between memory-intensive search restricted to fixed variable ordering and inference beyond: 1. different direction of exploring a common search space (top down for search vs. bottom-up for inference); 2. different assumption of control strategy (depth-first for search and breadth-first for inference). We also show that those differences have no practical effect, except under the presence of determinism. Our analysis assumes a fixed variable ordering. Some of the conclusions may not hold for dynamic variable ordering.

Section 2 provides background. Section 3 compares VE with AO search. Section 4 addresses the effect of advanced algorithmic schemes and section 5 concludes.

## 2 BACKGROUND

### 2.1 GRAPHICAL MODELS

A graphical model is defined by a collection of functions, over a set of variables, conveying probabilistic or deterministic information, whose structure is captured by a graph.

DEFINITION **2.1 (graphical models)** *A* graphical model *is a 4-tuple* $\mathcal{M}=\langle X, D, F, \otimes \rangle$, *where: 1.* $X=\{X_1,\ldots,X_n\}$ *is a set of variables; 2.* $D=\{D_1,\ldots,D_n\}$ *is a set of finite domains of values; 3.* $F=\{f_1,\ldots,f_r\}$ *is a set of real-valued functions. The scope of function* $f_i$, *denoted* $\text{scope}(f_i) \subseteq X$, *is the set of arguments of* $f_i$ *4.* $\otimes_i f_i \in \{\prod_i f_i, \sum_i f_i, \bowtie_i f_i\}$ *is a combination operator. The graphical model represents the combination of all its functions, namely the set:* $\otimes_{i=1}^r f_i$. *When the combination operator is irrelevant we denote* $\mathcal{M}$ *by* $\langle X, D, F \rangle$.

DEFINITION **2.2 (primal graph)** *The* primal graph *of a graphical model is an undirected graph that has variables as its vertices and edges connecting any two variables that appear in the scope of the same function.*

Two central graphical models are belief networks and constraint networks. A **belief network** $\mathcal{B} = \langle X, D, P \rangle$ is defined over a directed acyclic graph $G = (X, E)$ and its functions $P_i$ denotes conditional probability tables (CPTs), $P_i = \{P(X_i \mid pa_i)\}$, where $pa_i$ is the set of *parent* nodes pointing to $X_i$ in G. Common tasks are finding the posterior probability of some variables given the evidence, or finding the most probable assignment to all the variables given the evidence. A **constraint network** $\mathcal{R} = \langle X, D, C \rangle$ has a set of constraints $C = \{C_1, ..., C_t\}$ as its functions. Each constraint is a pair $C_i = (S_i, R_i)$, where $S_i \subseteq X$ is the scope of the relation $R_i$ defined over $S_i$, denoting the allowed combination of values. Common tasks are finding a solution and counting the number of solutions.

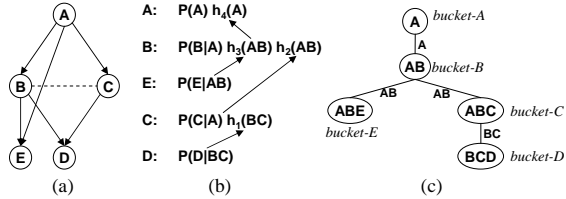

Figure 1: Execution of Variable Elimination

We assume that the domains of functions include a zero element, "0". Combining (*e.g.*, multiplying) anything with "0" yields a "0". The "0" value expresses inconsistent tuples. This is a primary concept in constraint networks but can also be defined relative to a graphical model as follows. Each function $f_i$ expresses an implicit flat constraint which is a relation $R_i$ over its scope that includes all and only the consistent tuples, namely those that are not mapped to "0". In this paper, a constraint network refers also to the flat constraints that can be extracted from any graphical model. When all the assignments are consistent we say that the graphical model is strictly positive. A partial assignment is consistent if none of its functions evaluate to zero. A solution is a consistent assignment to all the variables.

We assume the usual definitions of *induced graphs*, *induced width*, *tree-width* and *path-width* [9, 1].

## 2.2 INFERENCE BY VARIABLE ELIMINATION

Variable elimination algorithms [5, 8] are characteristic of inference methods. Consider a graphical model $G = \langle X, D, F \rangle$ and an ordering $d = (X_1, X_2, \ldots, X_n)$. The ordering $d$ dictates an elimination order for VE, from last to first. All functions in $F$ that contain $X_i$ and do not contain any $X_j, j > i$, are placed in the *bucket of* $X_i$. Buckets are processed from $X_n$ to $X_1$ by eliminating the bucket variable (combining all functions and removing the variable by a marginalization) and placing the resulting function (also called *message*) in the bucket of its latest variable in $d$. This VE procedure also constructs a bucket tree, by linking each bucket $X_i$ to the one where the resulting function generated in bucket $X_i$ is placed, which is called the parent of $X_i$.

**Example 2.1** *Figure 1a shows a belief network. Figure1b shows the execution of Variable Elimination along ordering* $d = (A, B, E, C, D)$. *The buckets are processed from D to A* [1]. *Figure 1c shows the bucket tree.*

## 2.3 AND/OR SEARCH SPACE

The usual way to do search consists of instantiating variables in turn (we only consider fixed variable ordering). In the simplest case this defines a search tree, whose nodes represent states in the space of partial assignments. A depth-first search (DFS) algorithm searching this space could run in linear space. If more memory is available, then some of the traversed nodes can be cached, and retrieved when "similar" nodes are encountered. The traditional search space does not capture the structure of the underlying graphical model. Introducing $AND$ nodes into the search space can capture the structure decomposing the problem into independent subproblems by conditioning on values [10, 12]. Since the size of the AND/OR tree may be exponentially smaller than the traditional search tree, any algorithm searching the AND/OR space enjoys a better computational bound. For more details see [4, 12]. A classical algorithm that explores the AND/OR search space is Recursive Conditioning [7]. Given a graphical model $\mathcal{M}$, its AND/OR search space is based on a *pseudo tree* [12]:

DEFINITION **2.3 (pseudo tree)** *Given an undirected graph* $G = (V, E)$, *a directed rooted tree* $T = (V, E')$ *defined on all its nodes is called* pseudo tree *if any arc of* $G$ *which is not included in* $E'$ *is a back-arc, namely it connects a node to an ancestor in* $T$.

### 2.3.1 AND/OR Search Tree

Given a graphical model $\mathcal{M} = \langle X, D, F \rangle$, its primal graph $G$ and a pseudo tree $T$ of $G$, the associated AND/OR search tree, denoted $S_T(\mathcal{M})$, has alternating levels of AND and OR nodes. The OR nodes are labeled $X_i$ and correspond to the variables. The AND nodes are labeled $\langle X_i, x_i \rangle$ and correspond to the values in the domains of the variables. The structure of the AND/OR search tree is based on the underlying backbone pseudo tree $T$. The root of the AND/OR search tree is an OR node labeled with the root of $T$.

The children of an OR node $X_i$ are AND nodes labeled with assignments $\langle X_i, x_i \rangle$ that are consistent with the assignments along the path from the root, $path(x_i) = (\langle X_1, x_1 \rangle, \langle X_2, x_2 \rangle, \ldots, \langle X_{i-1}, x_{i-1} \rangle)$. Consistency is well defined for constraint networks. For probabilistic networks it is defined relative to the underlying flat constraint network derived from the belief network. The children of an AND node $\langle X_i, x_i \rangle$ are OR nodes labeled with the children of variable $X_i$ in the pseudo tree $T$.

**Arc labeling** The arcs from $X_i$ to $\langle X_i, x_i \rangle$ are labeled with the appropriate combined values of the functions in $F$ that contain $X_i$ and have their scopes fully assigned. When the pseudo tree is a chain, the AND/OR search tree coincides with the regular OR search tree.

---
[1]This is a non-standard graphical representation, reversing the top down bucket processing described in earlier papers.

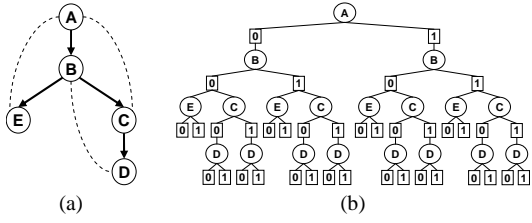

Figure 2: AND/OR search tree

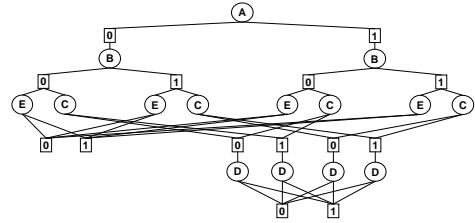

Figure 3: Context-minimal AND/OR search graph

**Example 2.2** *Consider again the belief network in Figure 1a. Figure 2a shows a pseudo tree of its primal graph, together with the back-arcs (dotted lines). Figure 2b shows the AND/OR search tree based on the pseudo tree, for binary $\{0, 1\}$ valued variables assuming positive functions. Arc labels are not included.*

Based on earlier work by [12, 4, 6, 7], it can be shown that:

THEOREM **2.3** *Given a graphical model $\mathcal{M}$ and a pseudo tree $T$, the size of the AND/OR search tree $S_T$ is $O(n \cdot \exp(m))$ where $m$ is the depth of $T$. A graphical model that has a tree-width $w^*$ has an AND/OR search tree whose size is $O(n \cdot \exp(w^* \cdot \log n))$.*

DEFINITION **2.4 (backtrack-free)** *An AND/OR search tree of a graphical model is* backtrack-free *iff all nodes that do not root a consistent solution are pruned.*

### 2.3.2 AND/OR Search Graph

The AND/OR search tree may contain nodes that root identical subtrees. These are called *unifiable*. When unifiable nodes are merged, the search space becomes a graph. Its size becomes smaller at the expense of using additional memory when being traversed. When all unifiable nodes are merged, a computational intensive task, we get the unique *minimal AND/OR graph*. Some unifiable nodes can be identified based on their *contexts* [7] or *conflict sets* [9]. The context of an AND node $\langle X_i, x_i \rangle$ is defined as the set of ancestors of $X_i$ in the pseudo tree, including $X_i$, that are connected (in the induced primal graph) to descendants of $X_i$. It is easy to verify that the context of $X_i$ d-separates [14] the subproblem below $X_i$ from the rest of the network. The *context-minimal* AND/OR graph denoted $CM_T(\mathcal{M})$, is obtained by merging all the context unifiable AND nodes. When the graphical model is strictly positive, it yields the *full context-minimal graph*. The *backtrack-free context-minimal graph*, $BF\text{-}CM_T$, is the context-minimal graph where *all* inconsistent subtrees are pruned.

**Example 2.4** *Figure 3 shows the full context-minimal graph of the problem and pseudo tree from Figure 2.*

Based on earlier work by [4, 11], it can be shown that:

THEOREM **2.5 (size of minimal context graphs)** *Given a graphical model $\mathcal{M}$, a pseudo tree $T$ and $w$ the induced width of $G$ along the depth-first traversal of $T$,*
*1) The size of $CM_T(\mathcal{M})$ is $O(n \cdot k^w)$, when $k$ bounds the domain size.*
*2) The context-minimal AND/OR search graph (relative to all pseudo trees) is bounded exponentially by the tree-width, while the context-minimal OR search graph is bounded exponentially by the path-width.*

**Value function** A task over a graphical model (*e.g.*, belief updating, counting) induces a value function for each node in the AND/OR space. The algorithmic task is to compute the value of the root. This can be done recursively from leaves to root by any traversal scheme. When an AO traversal of the search space uses full caching based on context it actually traverses the context-minimal, $CM_T$, graph. It is this context minimal graph that allows comparing the execution of AO search against VE.

## 3 VE VS. AO SEARCH

VE's execution is uniquely defined by a bucket-tree, and since every bucket tree corresponds to a pseudo tree, and a pseudo tree uniquely defines the context-minimal AND/OR search graph, we can compare both schemes on this common search space. Furthermore, we choose the context-minimal AND/OR search graph (CM) because algorithms that traverse the full CM need memory which is comparable to that used by VE, namely, space exponential in the tree-width of their pseudo/bucket trees.

Algorithm AO denotes any traversal of the CM search graph, AO-DF is a depth-first traversal and AO-BF is a breadth-first traversal. We will compare VE and AO via the portion of this graph that they generate and by the order of node generation. The task's value computation performed during search adds only a constant factor.

We distinguish graphical models with or without determinism, namely, graphical models that have inconsistencies vs. those that have none. We compare *brute-force* versions of VE and AO, as well as versions enhanced by various known features. We assume that the task requires the examination of all solutions (*e.g.* belief updating, counting solutions).

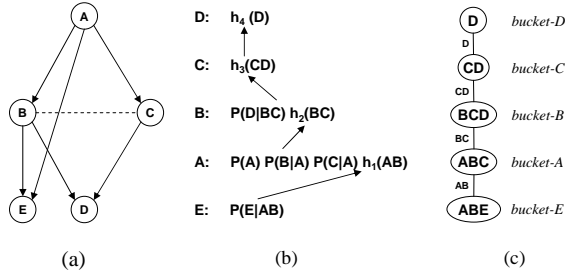

Figure 4: Variable Elimination

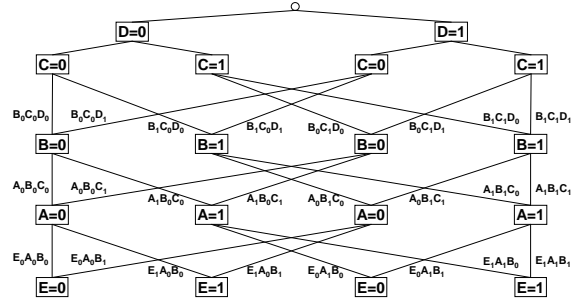

Figure 5: Context-minimal AND/OR search space

## 3.1 VE VS. AO WITH NO DETERMINISM

We start with the simplest case in which the graphical model contains no determinism and the bucket tree (pseudo tree) is a chain.

### 3.1.1 OR Search Spaces

Figure 4a shows a Bayesian network. Let's consider the ordering $d = (D, C, B, A, E)$ which has the tree-width $w(d) = w^* = 2$. Figure 4b shows the bucket-chain and a schematic application of VE along this ordering (the bucket of E is processed first, and the bucket of D last). The buckets include the initial CPTs and the functions that are generated and sent (as messages) during the processing. Figure 4c shows the bucket tree.

If we use the chain bucket tree as pseudo tree for the AND/OR search along $d$, we get the *full CM graph* given in Figure 5. Since this is an OR space, we can eliminate the OR levels as shown. Each level of the graph corresponds to one variable. The edges should be labeled with the product of the values of the functions that have just been instantiated on the current path. We note on the arc just the assignment to the relevant variables (*e.g.*, $B_1$ denotes $B = 1$). For example, the edges between C and B in the search graph are labeled with the function valuation on $(BCD)$, namely $P(D|B,C)$, where for each individual edge this function is instantiated as dictated on the arcs.

AO-DF computes the value (*e.g.*, updated belief) of the root node by generating and traversing the context-minimal graph in a depth-first manner and accumulating the partial value (*e.g.*, probabilities) using combination (products) and marginalization (summation). The first two paths generated by AO-DF are $(D_0, C_0, B_0, A_0, E_0)$ and $(D_0, C_0, B_0, A_0, E_1)$, which allow the first accumulation of value $h_1(A_0B_0) = P(E_0|A_0B_0) + P(E_1|A_0B_0)$. AO-DF subsequently generates the two paths $(D_0, C_0, B_0, A_1, E_0)$ and $(D_0, C_0, B_0, A_1, E_1)$ and accumulates the next partial value $h_1(A_1B_0) = P(E_0|A_1B_0) + P(E_1|A_1B_0)$. Subsequently it computes the summation $h_2(B_0C_0) = P(A_0) \cdot P(B_0|A_0) \cdot P(C_0|A_0) \cdot h_1(A_0B_0) + P(A_1) \cdot P(B_0|A_1) \cdot P(C_0|A_1) \cdot h_1(A_1B_0)$. Notice that due to caching each arc is generated and traversed just once (in each direction). For example when the partial path $(D_1, C_0, B_0)$ is created, it is recognized (via context) that the subtree below was already explored and its compiled value will be reused.

In contrast, VE generates the full context-minimal graph by layers, from the *bottom of the search graph up*, in a manner that can be viewed as dynamic programming or as breadth-first search on the explicated search graph. VE's execution can be viewed as first generating all the edges between E and A (in some order), and then all the edges between A and B (in some order), and so on up to the top. We can see that there are 8 edges between E and A. They correspond to the 8 tuples in the bucket of E (the function on $(ABE)$). There are 8 edges between A and B, corresponding to the 8 tuples in the bucket of A. And there are 8 edges between B and C, corresponding to the 8 tuples in the bucket of B. Similarly, 4 edges between C and D correspond to the 4 tuples in the bucket of C, and 2 edges between D and the rood correspond to the 2 tuples in the bucket of D.

Since the computation is performed from bottom to top, the nodes of A store the result of *eliminating* E (namely the function $h_1(AB)$ resulting by summing out $E$). There are 4 nodes labeled with A, corresponding to the 4 tuples in the message sent by VE from bucket of E to bucket of A (the message on $(AB)$). And so on, each level of nodes corresponds to the number of tuples in the message sent on the separator (the common variables) between two buckets.

### 3.1.2 AND/OR Graphs

The above correspondence between Variable Elimination and AND/OR search is also maintained in non-chain pseudo/bucket trees. We refer again to the example in Figures 1, 2 and 3 and assume belief updating. The bucket tree in Figure 1c has the same structure as the pseudo tree in Figure 2a. We will show that VE traverses the AND/OR search graph in Figure 3 bottom up, while AO-DF traverses the same graph in depth first manner, top down.

AO-DF first sums $h_3(A_0, B_0) = P(E_0|A_0, B_0) + P(E_1|A_0, B_0)$ and then goes depth first to $h_1(B_0, C_0) = P(D_0|B_0, C_0) + P(D_1|B_0, C_0)$ and $h_1(B_0, C_1) = P(D_0|B_0, C_1) + P(D_1|B_0, C_1)$. Then it computes

$h_2(A_0, B_0) = (P(C_0|A_0) \cdot h_1(B_0, C_0)) + (P(C_1|A_0) \cdot h_1(B_0, C_1))$. All the computation of AO-DF is precisely the same as the one performed in the buckets of VE. Namely, $h_1$ is computed in the bucket of $D$ and placed in the bucket of $C$. $h_2$ is computed in the bucket of $C$ and placed in the bucket of $B$, $h_3$ is computed in the bucket of $E$ and also placed in the bucket of $B$ and so on, as shown in Figure 1b. All this corresponds to traversing the AND/OR graph from leaves to root. Thus, both algorithms traverse the same graph, only the control strategy is different.

We can generalize both the OR and AND/OR examples,

THEOREM **3.1 (VE and AO-DF are identical)** *Given a graphical model having no determinism, and given the same bucket/pseudo tree VE applied to the bucket-tree is a (breadth-first) bottom-up search that will explore all the full CM search graph, while AO-DF is a depth-first top-down search that explores (and records) the full CM graph as well.*

**Breadth-first on AND/OR.** Since one important difference between AO search and VE is the order by which they explore the search space (top-down vs. bottom-up) we wish to remove this distinction and consider a VE-like algorithm that goes top-down. One obvious choice is breadth-first search, yielding AO-BF. That is, in Figure 3 we can process the layer of variable A first, then B, then E and C, and then D. General *breadth-first* or *best-first* search of AND/OR graphs for computing the optimal cost solution subtrees are well defined procedures. The process involves expanding all solution subtrees in layers of depth. Whenever a new node is generated and added to the search frontier the value of all relevant partial solution subtrees are updated. A well known Best-first version of AND/OR spaces is the AO* algorithm [13]. Algorithm AO-BF can be viewed as a top-down inference algorithm. We can now extend the comparison to AO-BF.

**Proposition 1 (VE and AO-BF are identical)** *Given a graphical model with no determinism and a bucket/pseudo tree, VE and AO-BF explore the same full CM graph, one bottom-up (VE) and the other top-down; both perform identical value computation.*

**Terminology for algorithms' comparison.** Let $A$ and $B$ be two algorithms over graphical models, whose performance is determined by an underlying bucket/pseudo tree.

DEFINITION **3.1 (comparison of algorithms)** *We say that: 1. algorithms $A$ and $B$ are identical if for every graphical model and when given the same bucket-tree they traverse an identical search space. Namely, every node is explored by $A$ iff it is explored by $B$; 2. $A$ is weakly better than $B$ if there exists a graphical model and a bucket-tree, for which $A$ explores a strict subset of the nodes explored by $B$; 3. $A$ is better than $B$ if $A$ is weakly better than $B$ but $B$ is not weakly better than $A$; 4. The relation of "weakly-better" defines a partial order between algorithms. $A$ and $B$ are* not comparable *if they are not comparable w.r.t to the "weakly-better" partial order.*

Clearly, any two algorithms for graphical models are either 1. identical, 2. one is better than the other, or 3. they are not comparable. We can now summarize our observations so far using the new terminology.

THEOREM **3.2** *For a graphical model having no determinism AO-DF, AO-BF and VE are identical.*

Note that our terminology captures the time complexity but may not capture the space complexity, as we show next.

### 3.1.3 Space Complexity

To make the complete correspondence between VE and AO search, we can look not only at the computational effort, but also at the space required. Both VE and AO search traverse the context minimal graph, but they may require different amounts of memory to do so. So, we can distinguish between the portion of the graph that is traversed and the portion that should be recorded and maintained. If the whole graph is recorded, then the space is $O(n \cdot \exp(w^*))$, which we will call the base case.

**VE can forget layers** Sometimes, the task to be solved can allow VE to use less space by deallocating the memory for messages that are not necessary anymore. Forgetting previously traversed layers of the graph is a well known property of dynamic programming. In such a case, the space complexity for VE becomes $O(d_{BT} \cdot \exp(w*))$, where $d_{BT}$ is the *depth* of the bucket tree (assuming constant degree in the bucket tree). In most cases, the above bound is not tight. If the bucket tree is a chain, then $d_{BT} = n$, but forgetting layers yields an $O(n)$ improvement over the base case. AO-DF cannot take advantage of this property of VE. It is easy to construct examples where the bucket tree is a chain, for which VE requires $O(n)$ less space than AO-DF.

**AO dead caches** The straightforward way of caching is to have a table for each variable, recording its context. However, some tables might never get cache hits. We call these *dead-caches*. In the AND/OR search graph, dead-caches appear at nodes that have only one incoming arc. AO search needs to record only nodes that are likely to have additional incoming arcs, and these nodes can be determined by inspection from the pseudo tree. Namely, if the context of a node includes that of its parent, then AO need not store anything for that node, because it would be a dead-cache.

In some cases, VE can also take advantage of dead caches. If the dead caches appear along a chain in the pseudo tree, then avoiding the storage of dead-caches in AO corresponds to collapsing the subsumed neighboring buckets in the bucket tree. This results in having cache tables of the size of the separators, rather than the cliques. The time savings are within a constant factor from the complexity of solving the largest clique, but the space complexity can be reduced from exponential in the size of the maximal cique to exponential in the maximal separator.

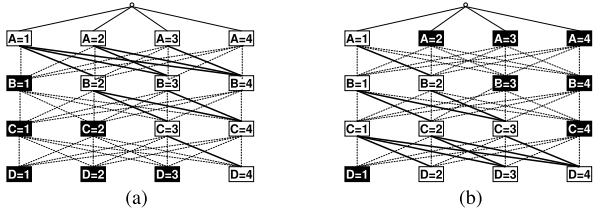

Figure 6: CM graphs with determinism: a) AO; b) VE

However, if the variables having dead caches form connected components that are subtrees (rather than chains) in the pseudo tree, then the space savings of AO cannot be achieved by VE. Consider the following example:

**Example 3.3** *Let variables $\{X_1, \ldots, X_n\}$ be divided in three sets: $\mathcal{A} = \{X_1, \ldots, X_{\frac{n}{3}}\}$, $\mathcal{B} = \{X_{\frac{n}{3}+1}, \ldots, X_{\frac{2n}{3}}\}$ and $\mathcal{C} = \{X_{\frac{2n}{3}+1}, \ldots, X_n\}$. There are two cliques on $\mathcal{A} \cup \mathcal{B}$ and $\mathcal{A} \cup \mathcal{C}$ defined by all possibile binary functions over variables in those respective cliques. The input is therefore $O(n^2)$. Consider the bucket tree (pseudo tree) defined by the ordering $d = (X_1, \ldots, X_n)$, where $X_n$ is eliminated first by VE. In this pseudo tree, all the caches are dead, and as a result the AO search graph coincides with the AO search tree. Therefore, AO can solve the problem using space $O(\frac{2n}{3})$. VE can collapse some neighboring buckets (for variables in $\mathcal{B}$ and $\mathcal{C}$), but needs to store at least one message on the variables in $\mathcal{A}$, which yields space complexity $O(\exp(\frac{n}{3}))$. In this example, AO and VE have the same time complexity, but AO uses space linear in the number of variables while VE needs space exponential in the number of variables (and exponential in the input too).*

The above observation is similar to the known properties of depth-first vs. breadth-first search in general. When the search space is close to a tree, the benefit from the inherent memory use of breadth-first search is nil.

### 3.2 VE VS. AO WITH DETERMINISM

When the graphical model contains determinism the AND/OR trees and graphs are dependant not only on the primal graph but also on the (flat) constraints, namely on the consistency and inconsistency of certain relationships (no-good tuples) in each relation. In such cases AO and VE, may explore different portions of the context-minimal graphs because the order of variables plays a central role, dictating where the determinism reveals itself.

**Example 3.4** *Let's consider a problem on four variables: $A, B, C, D$, each having the domain $\{1, 2, 3, 4\}$, and the constraints $A < B$, $B < C$ and $C < D$. The primal graph of the problem is a chain. Let's consider the natural ordering from A to D, which gives rise to a chain pseudo tree (and bucket-tree) rooted at A. Figure 6a shows the full CM graph with determinism generated by AO search, and Figure 6b the graph generated and traversed by VE in reverse order. The thick lines and the white nodes are the ones traversed. The dotted lines and the black nodes are not explored (when VE is executed from D, the constraint between D and C implies that $C = 4$ is pruned, and therefore not further explored). Note that the intersection of the graphs explored by both algorithms is the backtrack-free AND/OR context graph, corresponding to the unique solution (A=1,B=2,C=3,D=4).*

As we saw in the example, AO and VE explore different parts of the inconsistent portion of the full CM. Therefore, in the presence of determinism, AO-DF and AO-BF are both un-comparable to VE, as they differ in the direction they explore the CM space.

THEOREM **3.5** *Given a graphical model with determinism, then AO-DF and AO-BF are identical and both are un-comparable to VE.*

This observation is in contrast with claims of superiority of one scheme or the other [3], at least for the case when variable ordering is fixed and no advanced constraint propagation schemes are used and assuming no exploitation of context independence.

## 4 ALGORITHMIC ADVANCES AND THEIR EFFECT

So far we compared brute-force VE to brute-force AO search. We will now consider the impact of some enhancements on this relationship. Clearly, both VE and AO explore the portion of the context-minimal graph that is backtrack-free. Thus they can differ only on the portion that is included in full CM and not in the backtrack-free CM. Indeed, constraint propagation, backjumping and nogood recording just reduce the exploration of that portion of the graph that is *inconsistent*. Here we compare those schemes against bare VE and against VE augmented with similar enhancements whenever relevant.

### 4.1 VE VS. AO WITH LOOK-AHEAD

In the presence of determinism AO-DF and AO-BF can naturally accommodate look-ahead schemes which may avoid parts of the context-minimal search graph using some level of constraint propagation. It is easy to compare AO-BF against AO-DF when both use the same look-ahead because the notion of constraint propagation as look-ahead is well defined for search and because both algorithms explore the search space top down. Not surprisingly when both algorithms have the same level of look-ahead propagation, they explore an identical search space.

We can also augment VE with look-ahead constraint propagation (*e.g.*, unit resolution, arc consistency), yielding VE-LAH as follows. Once VE-LAH processes a single bucket, it then applies constraint propagation as dictated by the look-ahead propagation scheme (bottom-up), then continues with the next bucket applied over the modified set of functions and so on. We can show that:

THEOREM **4.1** *Given a graphical model with determinism and given a look-ahead propagation scheme, LAH,*
*1. AO-DF-LAH and AO-BF-LAH are identical.*
*2. VE and VE-LAH are each un-comparable with each of AO-DF-LAH and AO-BF-LAH.*

## 4.2 GRAPH-BASED NO-GOOD LEARNING

AO search can be augmented with no-good learning [9]. Graph-based no-good learning means recording that some nodes are inconsistent based on their context. This is automatically accomplished when we explore the CM graph which actually amounts to recording no-goods and goods by their contexts. Therefore AO-DF is identical to AO-BF and both already exploit no-goods, we get that (AO-NG denotes AO with graph-based no-good learning):

THEOREM **4.2** *For every graphical model the relationship between AO-NG and VE is the same as the relationship between AO (Depth-first or breadth-first) and VE.*

**Combined no-goods and look-ahead.** No-goods that are generated during search can also participate in the constraint propagation of the look-ahead and strengthen the ability to prune the search-space further. The graphical model itself is modified during search and this affects the rest of the look-ahead. It is interesting to note that AO-BF is not able to simulate the same pruned search space as AO-DF in this case because of its breadth-first manner. While AO-DF can discover deep no-goods due to its depth-first nature, AO-BF has no access to such deep no-goods and cannot use them within a constraint propagation scheme in shallower levels. However, even when AO exploits no-goods within its look-ahead propagation scheme, VE and AO remain un-comparable. Any example that does not allow effective no-good learning can illustrate this.

**Example 4.3** *Consider a constraint problem over $n$ variables. Variables $X_1, \ldots, X_{n-1}$ have the domain $\{1, 2, \ldots, n-2, *\}$, made of n-2 integer values and a special $*$ value. Between any pair of the $n-1$ variables there is a not-equal constraint between the integers and equality between stars. There is an additional variable $X_n$ which has a constraint with each variable, whose values are consistent only with the $*$ of the other n-1 variables. Clearly if the ordering is $d = (X_1, \ldots, X_{n-1}, X_n)$, AO may search*

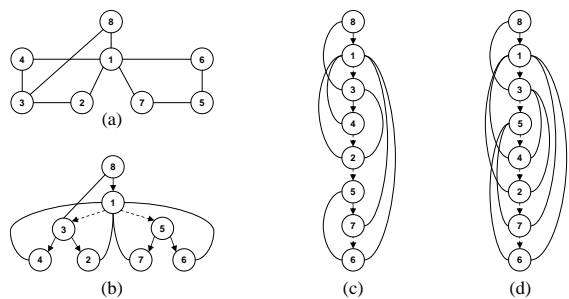

Figure 7: GBJ vs. AND/OR search

*all the exponential search space over the first $n-1$ variables (the inconsistent portion) before it reaches the $*$ of the $n-th$ variable. On the other hand, if we apply VE starting from the $n-th$ variable, we will reveal the only solution immediately. No constraint propagation, nor no-good learning can help any AO search in this case.*

THEOREM **4.4** *Given a graphical model with determinism and a particular look-ahead propagation scheme LAH:*
*1. AO-DF-LAH-NG is better than AO-BF-LAH-NG.*
*2. VE and AO-DF-LAH-NG are not comparable.*

## 4.3 GRAPH-BASED BACKJUMPING

Backjumping algorithms [9] are backtracking search applied to the OR space, which uses the problem structure to jump back from a dead-end as far back as possible. In *graph-based backjumping* (GBJ) each variable maintains a graph-based induced ancestor set which ensures that no solutions are missed by jumping back to its deepest variable.

**DFS orderings.** If the ordering of the OR space is a DFS ordering of the primal graph, it is known [9] that all the backjumps are from a variable to its DFS parent. This means that *naive AO-DF* automatically incorporates GBJ jumping-back character.

**Pseudo tree orderings.** In the case of pseudo tree orderings that are not DFS-trees, there is a slight difference between OR-GBJ and AO-DF and GBJ may sometime perform deeper backjumps than those implicitly done by AO. Figure 7a shows a probabilistic model, 7b a pseudo tree and 7c a chain driving the OR search (top down). If a deadend is encountered at variable 3, GBJ retreats to 8 (see 7c), while naive AO-DF retreats to 1, the pseudo tree parent. When the deadend is encountered at 2, both algorithms backtrack to 3 and then to 1. Therefore, in such cases, augmenting AO with GBJ can provide additional pruning on top of the AND/OR structure. In other words, GBJ on OR space along a pseudo tree is never stronger than GBJ on AND/OR and it is sometimes weaker.

GBJ can be applied using an arbitrary order $d$ for the OR space. The ordering $d$ can be used to generate a pseudo tree. In this case, however, to mimic GBJ on $d$, the AO traversal will be controlled by $d$. In Figure 7d we show an arbitrary order $d = (8, 1, 3, 5, 4, 2, 7, 6)$ which generates the pseudo tree in 7b. When AO search reaches 3, it goes in a breadth first manner to 5, according to $d$. It can be shown that GBJ in order $d$ on OR space corresponds to the GBJ-based AND/OR search based on the associated pseudo tree. All the backjumps have a one to one correspondence.

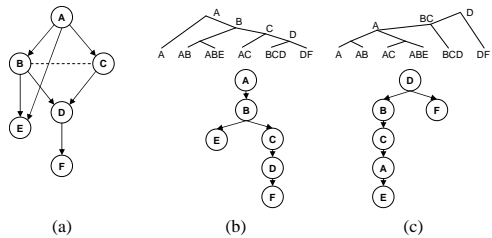

Figure 8: RC d-trees and AND/OR pseudo trees

Since VE is not comparable with AO-DF, it is also uncomparable with AO-DF-GBJ. Note that backjumping is not relevant to AO-BF or VE. In summary,

THEOREM **4.5** *1. When the pseudo tree is a DFS tree AO-DF is identical to AO-DF-GBJ. This is also true when the AND/OR search **tree** is explored (rather than the CM-**graph**). 2. AO-DF-GBJ is superior to AO-DF for general pseudo trees. 3. VE is not comparable to AO-DF-GBJ.*

### 4.4 RECURSIVE CONDITIONING AND VALUE ELIMINATION

**Recursive Conditioning (RC)** [7] defined for belief networks is based on the divide and conquer paradigm. RC instantiates variables with the purpose of breaking the network into independent subproblems, on which it can recurs using the same technique. The computation is driven by a data-structure called *dtree*, which is a full binary tree, the leaves of which correspond to the network CPTs.

It can be shown that RC explores an AND/OR space. Consider the example in Figure 8, which shows: (a) a belief network; (b) and (c), two dtrees and the corresponding pseudo trees for the AND/OR search. It can also be shown that the context of the nodes in RC is identical to that in AND/OR and therefore equivalent caching schemes can be used.

**Value Elimination** [3] is a recently developed algorithm for Bayesian inference. It was already explained in [3] that, under static variable ordering, there is a strong relation between Value Elimination and VE. We can therefore derive that Value Elimination also explores an AND/OR space under static variable orderings.

## 5 SUMMARY AND CONCLUSIONS

The paper compares search and inference in graphical models through the new framework of AND/OR search spaces. We show that there is no principled difference between memory-intensive search with fixed variable ordering and inference beyond: 1. different direction of exploring a common search space (top down for search vs. bottom-up for inference); 2. different assumption of control strategy (depth-first for search and breadth-first for inference). We also show that those differences occur only in the presence of determinism. We show the relationship between algorithms such as graph-based backjumping and no-good learning [9], as well as Recursive Conditioning [7] and Value Elimination [2] within the AND/OR search space. AND/OR search spaces can also accommodate dynamic variable and value ordering which can affect algorithmic efficiency significantly. Variable Elimination and general inference methods however require static variable ordering. This issue will be addressed in future work.

### Acknowledgments

This work was supported in part by the NSF grant IIS-0412854 and the MURI ONR award N00014-00-1-0617.